\documentclass[sigconf]{acmart}
\usepackage{multirow}
\usepackage{multicol}
\usepackage{subcaption}
\usepackage{color}
\usepackage{colortbl}
\usepackage{paralist}

\AtBeginDocument{%
  \providecommand\BibTeX{{%
    \normalfont B\kern-0.5em{\scshape i\kern-0.25em b}\kern-0.8em\TeX}}}

\setcopyright{acmcopyright}
\copyrightyear{2018}
\acmYear{2018}
\acmDOI{XXXXXXX.XXXXXXX}
\copyrightyear{2023}
\acmYear{2023}

\acmConference[Conference acronym 'XX]{Make sure to enter the correct
  conference title from your rights confirmation emai}{June 03--05,
  2018}{Woodstock, NY}
\acmPrice{15.00}
\acmISBN{978-1-4503-XXXX-X/18/06}

\begin{document}

\title{Diversity matters: Robustness of\\ bias measurements in Wikidata}

\author{Paramita Das}
\email{paramita.das@iitkgp.ac.in}
\affiliation{%
  \institution{Dept. of CSE, IIT Kharagpur}
  \country{India}}

\author{Sai Keerthana Karnam}
\email{karnamsaikeerthana@iitkgp.ac.in}
\affiliation{%
  \institution{Dept. of CSE, IIT Kharagpur}
  \country{India}}

\author{Anirban Panda}
\email{hadesanirban666@iitkgp.ac.in}
\affiliation{%
  \institution{Dept. of ISE, IIT Kharagpur}
  \country{India}}

\author{Bhanu Prakash Reddy Guda}
\email{bguda@cs.cmu.edu}
\affiliation{%
  \institution{Carnegie Mellon University}
  \country{USA}}
  
\author{Soumya Sarkar}
\email{soumyasarkar@microsoft.com}
\affiliation{%
  \institution{Microsoft IDC}
  \country{India}}

\author{Animesh Mukherjee}
\email{animeshm@cse.iitkgp.ac.in}
\affiliation{%
  \institution{Dept. of CSE, IIT Kharagpur}
  \country{India}}

\renewcommand{\shortauthors}{P. Das, et al.}

\begin{abstract}
With the widespread use of knowledge graphs (KG) in various automated AI systems and applications, it is very important to ensure that information retrieval algorithms leveraging them are free from societal biases. Previous works have depicted biases that persist in KGs, as well as employed several metrics for measuring the biases. However, such studies lack in the systematic exploration of the sensitivity of the bias measurements, through varying sources of data, or the embedding algorithms used. To address this research gap, in this work, we present a holistic analysis of bias measurement on the knowledge graph. First, we attempt to reveal data biases that surface in Wikidata for thirteen different demographics selected from seven continents. Next, we attempt to unfold the variance in the detection of biases by two different knowledge graph embedding algorithms - TransE and ComplEx. We conduct our extensive experiments on a large number of occupations sampled from the thirteen demographics with respect to the sensitive attribute, i.e., gender. Our results show that the inherent data bias that persists in KG can be altered by specific algorithm bias as incorporated by KG embedding learning algorithms. Further, we show that the choice of the state-of-the-art KG embedding algorithm has a strong impact on the ranking of biased occupations irrespective of gender. In particular, we find that the embedding algorithm ComplEx is more robust to the choice of demographics compared to TransE. Subsequently, we observe that the similarity of the biased occupations across demographics is minimal which reflects the socio-cultural differences around the globe. This is often overlooked by most of the coarse-grained approaches working at the aggregate level. We believe that this  full-scale audit of the bias measurement pipeline will raise awareness among the community while deriving insights related to design choices of data and algorithms both and refrain itself from the popular dogma of ``one-size-fits-all''.
\end{abstract}





\maketitle

\begin{sloppypar}
\section{Introduction}
With the rapid expansion of content available on the web, \textit{Knowledge Graph} (KG) is being widely used by tech giants to assimilate factual information in a structured format that can be used in many industrial applications. Academic research on KGs has been gaining significant interest for exploring knowledge in the light of representation of large-scale graph-structure data, reasoning, and application in a variety of automated tasks - recommendation~\cite{wang2021learning}, chatbot question answering~\cite{huang2019knowledge}, language modelling~\cite{Peters2019KnowledgeEC,zhang2019ernie} etc. In almost every downstream task of machine learning, knowledge extracted from knowledge graphs is assumed as the gold standard that establishes the correctness of the typical system. Like DBpedia\footnote{\url{https://www.dbpedia.org/resources/knowledge-graphs/}}, Google Knowledge Graph\footnote{\url{https://developers.google.com/knowledge-graph}}, Wikidata\footnote{\url{https://www.wikidata.org/wiki/Wikidata:Main_Page}} is an open collaborative knowledge base that was created to serve as the hub of structured linked data to all Wikimedia projects as well as to external digital tools and bots. Over the years, based on the efforts of human editors and automated software, Wikidata has become a giant knowledge base of over 99 million entities in multiple languages\footnote{\url{https://www.wikidata.org/wiki/Wikidata:Statistics}}. However, while knowledge graphs are evolving continually, the user-generated content of knowledge graphs mediates societal disparities in many downstream applications. For example, authors in \cite{zhang2021quantifying} have found that less than 22\% of Wikidata items represent people who are women and thus present a severe gender disparity in its content. 

\noindent In fact, a wide range of societal and human biases can be attributed to the curation of KGs. The entities and relationships in a typical KG are accumulated in a (semi) automatic way~\cite{demartini2019implicit,janowicz2018debiasing}, which may result in gathering biased knowledge from the open text corpus of the web. Furthermore, algorithms used to sample, aggregate, and process knowledge can incorporate biases into KGs. In handshake with the proliferation in embedding learning methods~\cite{dai2020survey,ji2021survey}, recent works have established the anecdotal presence of societal biases in KG data and how they are being mirrored by state-of-the-art KG embedding algorithms~\cite{fisher2020measuring,bourli2020bias}. Biases encoded in KGs and knowledge graph embeddings (KGEs) have a negative effect on society as well as the underlying automation systems that leverage the knowledge extracted from KGs in building the system. To tackle this issue, researchers have come up with coherent frameworks of bias measurement and debiasing them further~\cite{mlg2020_39}. Previous works~\cite{bourli2020bias,fisher2020measuring} on identifying biases are mostly focused on specific KGs  or typical KG embedding learning algorithms but their further comparison in the context of sensitive attributes and differences of societal constraints is missing in the literature. There could be such scenarios in which biases incurred from a sensitive attribute can vary across socio-economic, socio-cultural, and geographical boundaries fixing the knowledge graph upfront. For instance, we have found evidence in our experiment that the occupation ``activist'' is a female-dominated occupation in the middle-east demographics while it is a gender-neutral occupation in the western world considering the fact that \textit{gender} is a sensitive attribute.

\noindent\textbf{Our contributions}: In this work, we hypothesize that the choice of geo-social data, i.e., knowledge graphs pertaining to different demographics and KG embedding representation algorithms has a significant influence on the behavior of bias measurement in KGs and may lead to notable variability of biases depending on design choices. To demonstrate the variance of societal biases that exist in KGs, we introduce an empirical data-driven analysis on Wikidata. Among a broad set of sensitive attributes, gender is selected for our analysis since it leads to biases. The stereotypical observable that we consider here is professions or the dominated occupations by a specific gender. We assumed that the base dataset in our experiment itself contains biases resulting from cultural differences (aka \textit{data bias}). Note that such biases are inherent and cannot be therefore controlled in a crowd-sourced system. In this context, we have the following objectives-
\begin{compactitem}
    \item As the first objective, our point of interest is to check whether \textit{algorithmic biases} creep in when we build embedding from this base dataset. The hypothesis is that if they do then the data bias is ought to be altered.
    \item The second objective of this work is to audit the existence of gender bias (based on gender division - male and female) across two important orthogonal axes - (i) choice of training data modality and (ii) embedding learning algorithm.
\end{compactitem}
\noindent For this purpose, we collect a large number of Wikidata entities and relations from 13 different demographics around the globe - Arabia, Australia, Argentina, Brazil, France, Germany, India, Japan, Kenya, Russia, South Africa, United Kingdom, and the United States of America - ensuring representation from across the different continents. We conduct our experiments using two important knowledge graph embedding algorithms - TransE~\cite{bordes2013translating} and ComplEx~\cite{trouillon2016complex}. 
Our hypothesis is confirmed by our analysis; below we list the contributions of this work.
\begin{compactitem}
    \item We observe that the inherent data bias present in the base dataset is indeed revised by the embedding learning algorithm  demonstrating the presence of algorithmic biases.  
    \item With respect to a specific bias measurement metric (explained in section~\ref{sec:kge_metric}), the ranked lists of female/male dominated occupations are highly sensitive to both the embedding algorithm -- TransE and ComplEx and the underlying data demography. See the Appendix for results obtained by a third algorithm -- DistMult~\cite{yang2014embedding} which shows that our observations are generalized across different embedding learning algorithms.
\end{compactitem}
We believe that these are of serious concern since the conclusions drawn from different studies on KGs as well as the performance of the downstream applications dependent on KGs would vary based on the embedding algorithm and/or the data variance.

\noindent The paper is structured as follows. We have reviewed related literature in section~\ref{sec:related_work}. In section~\ref{sec:background}, we have discussed some preliminary ideas which are related to our work. A detailed description of our dataset is elaborated in section~\ref{sec:dataset}. In section~\ref{sec:experiment}, we lay out our experimental setup, followed by the key findings of our work in section~\ref{sec:result}. Finally, we conclude with some important notes and future scope in section~\ref{sec:discussion} and \ref{sec:conclusion} respectively.

\noindent All the codes, and data of the work are made available\footnote{\url{https://github.com/paramita08/Wikidata_Bias_WebSci_2023}} for reproducible research. 

\section{Related work} \label{sec:related_work}
Representation learning on KGs is a highly active direction in research, with numerous novel Knowledge Graph Embedding (KGE) algorithms being porposed recently, including \textit{TransE}~\cite{bordes2013translating}, \textit{TransD}~\cite{ji2015knowledge},
\textit{TransH}~\cite{wang2014knowledge}, \textit{RESCAL}~\cite{nickel2011three}, \textit{DistMult}~\cite{yang2014embedding}, \textit{HolE}~\cite{nickel2016holographic}, \textit{CrossE}~\cite{zhang2019interaction}, \textit{ComplEx}~\cite{trouillon2016complex} etc. Simultaneously with the recent progress in neural network approaches, several methods have been proposed using convolution kernels, among which \textit{ConvKB}~\cite{nguyen_2018}, \textit{ConvE}~\cite{dettmers2018convolutional}, \textit{HypER}~\cite{balavzevic2019hypernetwork} are important graph embedding methods. 
As often knowledge graphs are projected as collaborative repositories, it is quite understandable that knowledge graphs would be subjected to human perception and cognitive biases. Such social biases are reciprocated in terms of the distribution of entities and relations as well as get embedded in knowledge graph representations. Recent work~\cite{demartini2019implicit} on finding data biases in collaboratively constructed knowledge graphs, especially in Wikidata investigates how paid crowd-sourcing can be used to understand contributors' implicit bias. Specifically, the authors recruited crowd workers to verify controversial facts and demonstrated the benefits of surfacing bias information to end users of applications rather than eliminating them from the knowledge graph. Authors in \cite{zhang2021quantifying} find under representation of content about women as compared to men in Wikidata. Following the general predominance of the male population around the globe, Wikidata editors add many male-dominated occupations. However, Wikidata is no more biased than the real world; it mirrors the existing gender gap in our society. Similar work \cite{shaik2021analyzing} investigates the presence of race and citizenship bias in Wikidata. The study reveals that white individuals and those with citizenship in Europe and North America are over represented in comparison to the rest of the world. In similar lines, there exists research works~\cite{zagovora2017weitergeleitet}, attributed to Wikipedia, a sister project of Wikidata in which authors tried to reveal how imbalanced the gender presentation is on the occupations pages of
Wikipedia. Authors in~\cite{hollink2018using} studied gender differences in various Wikipedia language (European) editions with respect to the coverage of the
Members of the European Parliament (MEP). For further investigation, they inspected differences in the content of Wikidata entries of male and female MEPs and found gender imbalance across nationality. Besides the representation biases, researchers intended to characterize and mitigate the inference biases arising out of sensitive attributes, such as gender, ethnicity, religion etc. to finally make the KG embedding algorithms bias-free. Several works~\cite{bourli2020bias,fisher2020measuring} have successfully conducted experiments to show that harmful biases are penetrating societal spheres. Authors~\cite{keidar2021towards} in a recent work have suspected human assumptions related to the choice of sensitive relations. As a result, they proposed a framework that is capable of identifying biased attributes automatically based on some metrics. In an orthogonal direction, researchers are trying to invent various methodologies to mitigate biases from knowledge graphs which in turn will be helpful in designing a bias-free automated system for different machine learning and NLP tasks. As a useful solution, Arduini et al. \cite{mlg2020_39} developed a debiasing method based on adversarial learning that modifies the embedding by filtering
out sensitive information but preserving all the other
relevant information.  In~\cite{fisher2020debiasing} the authors presented a novel approach in which biased embeddings are trained to be neutral with respect to some sensitive attributes, such as gender-based on adversarial loss, and later users are allowed to add sensitive information back to the system on demand. This method has been proven to be significantly faster and more accurate than previous approaches~\cite{bose2019compositional} of debiasing knowledge graph embedding.

\par \noindent We take a step forward in this paper and tried to solve the following question– given a particular metric to measure bias score and a sensitive attribute (gender: female/male in this case), how (dis)similar are the bias score based ranked lists of a stereotypical observable (occupation in this case) obtained by varying (i) the KGE algorithm, and (ii) the demographic property of the underlying data. Ideally, the similarity between the two ranked lists on the basis of above mentioned two criteria should be high so that the effect on the downstream tasks and the conclusions drawn thereof remain same irrespective of the KGE algorithm used and the demographic variance of the data. Our observations are in stark contrast with this ideal scenario; the ranked lists are significantly diverse with very low overlap among them. We believe that this calls for immediate attention of the researchers and practitioners to deep dive into the proposed methodologies related to the biases in KGs.

\section{Background} \label{sec:background}
Formally, a knowledge graph is structured as a graph abstraction with nodes representing entities from a given domain and edges corresponding to relations between the entities. Ideally, facts in the knowledge graph are described as labeled triple format, such as $<h,r,t>$ where $h$ and $t$ are the head and tail entities respectively and $r$ denotes the relation between $h$ and $t$. For example, the knowledge triple \textit{$<$Joe Biden, born\_in, US$>$} expresses the fact that \textit{Joe Biden} was born in \textit{US}. Among the existent large number of knowledge graph embedding algorithms aimed at learning dense representations of the knowledge graphs, we have used two popular KG embedding algorithms, namely TransE and ComplEx in our experiment for simplicity and scalability of use.

\begin{enumerate}
    \item TransE~\cite{bordes2013translating}: The most elementary approach in KG embedding is the use of \textit{translational models} that assume a geometric perspective in which relation embeddings translate subject entity and relation to object entity in the low-dimensional space. The loss function $f_{TransE}$ is defined as the \textit{$L_{1}$} or \textit{$L_{2}$} norm  between the embedding of the tail and the embedding of the head plus the embedding of the relation as follows - $f_{TransE} = - ||h + r - t||_{n}$.

    \item ComplEx~\cite{trouillon2016complex}: This algorithm relies on \textit{tensor decomposition method} and each entity and relation is assumed as a complex vector (i.e., a vector containing complex numbers) of dimension $d$. The loss function of ComplEx is denoted as follows - $f_{ComplEx} = Re(<r,h,\overline t>)$. Here, $Re$ denotes the real vector component of the embedding generated in the complex space. 
\end{enumerate}

\subsection{Measuring data bias in KG}

To understand how biases are encoded in the knowledge triples of Wikidata, we tried to measure the data bias in our dataset of 13 demographics collected from Wikidata. A summary of the dataset is tabulated in Table~\ref{tab:datset_description} in section~\ref{sec:dataset}. For measuring the data bias that exists in KGs, we followed the bias measurement approach in KG data as proposed by Bourli~\cite{bourli2020bias} et al. We briefly describe the metric below.

\noindent \textbf{The metric:} Let us assume, a KG dataset contains of two types of triples primarily-- 
\textit{<human\_entity, has\_gender, gender>} and \textit{<human\_entity, has\_occupation, occupation>}. 
First, the metric computes the bias score $\theta$ for every occupations in the dataset. Formally, let us assume that for an occupation $o$, the number of male (female) entities that have occupation $o$ be $M_{o}$ ($F_{o}$) and $M$ ($F$) be the total number of male (female) entities that exist in the dataset. Now, given the occupation $o$, let $Pr(O = o | G = m)$ and $Pr(O = o | G = f)$ be the respective probabilities that male and female entities have the occupation $o$, where $O$, $G$, $m$, $f$ denote occupation, gender and its binary attributes - male and female respectively. For a given dataset, the probabilities and bias score are computed as follows -- (i) $Pr(O = o | G = m) = |M_{o}|/|M|$, (ii) $Pr(O = o | G = f) = |F_{o}|/|F|$ and (iii) $Pr(O = o | G = m) - Pr(O = o | G = f) = \theta$.

\noindent For a specific dataset, a threshold $t$ is estimated, $t$ being a positive value close to 0. For the occupation $o$, if the bias score $\theta > t$, the occupation is assumed to be male-biased, otherwise, if $\theta < -t$ then it is female-biased. Further, the occupations that have bias score within the range $[-t,t]$ are treated as neutral occupations. The threshold $t$ is selected from the distribution\footnote{The distribution is generated by plotting the count of neutral occupations for different values of $t$ and the optimal $t$ is selected as the value for which there is a sharp increase in the count of the neutral occupations.} of neutral occupations for different values of $t$. 

\subsection{Measuring biases in KG embedding}\label{sec:kge_metric}
Similar to the existence of social biases in word embeddings~\cite{bolukbasi2016man}, the biases in trained knowledge graph embeddings are prone to discriminative notions, for example, the occupation `banker' is more strongly tied to male than female entities. It results in many downstream applications skewed toward a particular section (e.g., based on gender) of society. Our objective is to investigate the influence of gender on the embedding method's outcome in predicting whether an occupation is male or female-biased or gender-neutral. In order to measure gender biases encoded in graph embedding of Wikidata, we followed one of the very popular bias measurement approaches, proposed by Fisher et al.~\cite{fisher2020measuring} which is briefly described below.

\noindent \textbf{Bias metric by Fisher et al.~\cite{fisher2020measuring}-}According to this metric, the first step is to update the initial embedding of a person in the knowledge graph to increase the representation of the male component in the person's embedding. This embedding could be a pretrained one or a representation uniformly sampled from an arbitrary distribution. The updation of embedding is achieved by providing the model with two batches of triples, $(e_{j}, r_{g}, e_{a})$ and $(e_{j}, r_{g}, e_{b})$, where $e_{j}$ is the embedding of the person $j$, $r_{g}$ is the embedding of the sensitive attribute (i.e., gender) and $e_{a}$, $e_{b}$ denotes the embedding of two primary values of the attribute gender, male and female respectively. The score function is denoted by $g(.)$ which takes the embeddings of a triple as input and outputs a score, denoting how likely this triple is to be correct. Next, we differentiate the score $m$ with respect to the embedding of person $j$, $e_{j}$, and take a step in the stochastic gradient descent algorithm with a learning rate $\alpha$ in order to maximize the score function as described in equation~\ref{eq:score_func} and equation~\ref{eq:update_func}.
\begin{equation}
    m(\theta) = g(e_{j} , r_{g}, e_{a}) - g(e_{j} , r_{g}, e_{b})
    \label{eq:score_func}
\end{equation}
\begin{equation}
    e^{\prime}_{j} = e_{j} + \alpha \frac{\delta m(\theta)}{\delta e_{j}}
    \label{eq:update_func}
\end{equation}
The second step is to calculate the difference of the scores given by the scoring function on the following triples $(e^{\prime}_{j}, r_{p}, e_{p})$ and $(e_{j}, r_{p}, e_{p})$ where $e^{\prime}_{j}$ is the updated embedding at the end of the updation process, $r_{p}$ is the embedding of the relation \textit{has\_occupation} and $e_{p}$ denotes the embedding of an occupation. Thus it assigns a bias score $b_{p}$ (i.e., the difference) to the occupation $p$. The above-mentioned steps are reiterated for all the human entities in an underlying knowledge graph comprised of two types of entities in general--\textit{<human\_entity, has\_gender, gender>} and \textit{<human\_entity, has\_occupation, occupation>}. Finally, the occupations that exist in the knowledge graph are ranked in descending order of their bias scores.

\noindent Formally, let us assume that we have two occupations $p_{1}$ and $p_{2}$ in our sample knowledge graph and the number of human entities is $N$. We calculate the bias scores $b_{p_{1}}$ and $b_{p_{2}}$ for the occupations $p_{1}$ and $p_{2}$ respectively following the steps as mentioned in the above bias measurement metric. In both cases, the bias score is computed over the set of all human entities $N$ in the knowledge graph. Now, we rank the bias scores and let us assume $b_{p_{1}} > {b_{p_{2}}}$. Hence, $p_{1}$ is ranked as a higher male-biased occupation than $p_{2}$. In other words, more the bias score, an occupation is closer to the gender line (i.e., the male part $e_{a}$ of the sensitive attribute gender than the female part $e_{b}$ in equation~\ref{eq:score_func}). The ranking of occupations in terms of descending order of bias score shows the likelihood of occupations ranked at the top to be more biased toward the male gender than the ones ranked later in the list. Similar to the above steps for obtaining the female-biased occupations, we update the embedding of human entities by taking the inverse of equation~\ref{eq:score_func}, and the occupations ranked at the top in the list are assumed as female-biased occupations. We have named the metric as embedding bias metric in later sections.

\noindent We note that other metrics like those proposed by Kediar et al.~\cite{keidar2021towards} have limitations. For instance, the metric Demographic Parity (DPD) as discussed by the authors relies more to the bias included in the ground truth data than KG embedding. Further, this metric is heavily dependent on the classifier used and the number of classes (i.e., the number of occupations according to our setting) and therefore has limited power of explanation. In contrast, the metric we have chosen here is the most generalized one and reflects the biases that are introduced by the embedding learning algorithms.

\section{Dataset description} \label{sec:dataset}

Wikidata is known as a free, open-source knowledge base that acts as knowledge storage for the structured data of sister Wikimedia projects such as Wikipedia, Wikivoyage, and others. Similar to other KGs, information is stored as facts or triples, containing a subject item, a property, and an object. Objects can be entities or literals such as a quantity, a string etc. The subject/object items
are denoted by URIs starting with `Q' (e.g. Q5284 for \textit{Bill Gates}) and properties are symbolized by URIs preceded by ’P’ (e.g., P19 for \textit{Place of birth}). We have extracted a specific set of triples from Wikidata based on a specific set of criteria and conducted experiments to find different trends of societal biases that exist in Wikidata.  

\subsection{Collection of demography dataset}
For our work, we downloaded the latest Wikidata dump\footnote{\url{https://dumps.wikimedia.org/wikidatawiki/entities/latest-all.json.bz2}} which is stored in a json format (bz2 format for compressed version). The dump consumes $\sim70$ GB space in compressed form. We first converted it into standard KGTK\footnote{\url{https://kgtk.readthedocs.io/en/latest/}} format for ease of data processing which generated three files: a node file ($\sim9.5$ GB), an edge file ($\sim189.5$ GB) and a qualifiers file ($\sim60.4$ GB). KGTK is a standard python library that facilitates easy manipulation of knowledge graphs. The node file contains the English labels, descriptions and aliases of Qnodes and Pnodes for examples, Qnode Q943 has label yellow, color as a description, and alias as the colour yellow/Y/FFFF00. The edge file contains the triplets in the knowledge graph along with the details of the tail-entity like language, entity-type etc. Next, we filtered out the triplets which are not having either a wiki QID as the head entity or a wiki PID as the relation. Thus we ended up with 1.32 billion triplets in the dataset with 93 million entities (QIDs) and 8763 relations (PIDs). Now, we aim to collect the triples corresponding to 13 demographics spanning over various continents which we have considered for our work - Arabia, Australia, Argentina, Brazil, France, Germany, India, Japan, Kenya, Russia, South Africa, United Kingdom,  and the United States of America.
The choice of the above demographics is motivated by the fact that the human entities covered by them are among the largest within their respective continents. This ensures large and diverse coverage of the dataset. The human entities belonging to a typical demography are found by computing the overlap between the entities that are humans and belonging to that specific demography. To find this connection, we first extract the head entities of the collected triplets which have the relation $P31$ (i.e., ``instance of'') and tail entity $Q5$ (i.e., ``human''). Further, the head entities of triplets must be connected with the relation $P27$ (i.e., ``country of citizenship'', here indicating the specific demography to which the entity belongs) and the tail entity as the corresponding QID of that country. Thus we gathered the entities belonging to each of the demographics. In addition, we considered all the outgoing edges from these human entities and collected all the triplets with the head-entity from the set of human entities we just obtained. This collects the gender and occupations information of all the entities. In this way, demography-specific knowledge graphs have been created. For example, for constructing the Arabia dataset we considered outgoing edges from the human entities belonging to any country in the Arabian Peninsula. A brief statistic of entities, triplets, humans, and occupations is shown in Table~\ref{tab:datset_description}.

\subsection{Formation of a giant knowledge graph}
To build a giant network joining the data of all the 13 demographics, entities of individual demography were combined together, which we denote by the set $E$. We constructed the network by considering the edges among these entities. For this purpose, we extracted all the triplets from the edge file with the head-entities and tail-entities that belonged to our set of entities $E$ and added to the graph. Finally, the giant subgraph of Wikidata that we constructed had 22,254,967 triplets, 2,228,594 entities, and 894 relations. We used English labels obtained from the node file and qwikidata\footnote{\url{https://qwikidata.readthedocs.io/en/stable/}} to find the labels of the QIDs and PIDs. For our experiments, we extracted the list of occupations that have at least one male and one female occurrence in a particular demography under consideration.

\begin{table}
\centering
\scriptsize
  \begin{tabular}{ |c||c|c|c|c|}
    \hline
    Demography & Triplets & Entities & Humans & occupations  \\
    \hline
    Arabia &1,63,730 & 18,813 &10,304 & 200\\\hline
    India &7,32,647 &85,287 &55,083 & 430\\\hline
    Japan & 26,73,572 & 2,09,637 &1,52,003 &683\\\hline
    Russia & 11,97,735& 1,00,770 & 55,548&676\\\hline
    Australia &9,53,677  & 99,257& 52,172&746\\\hline
    Kenya & 45,224& 6437 &3220 &100\\\hline
    South Africa & 2,34,490 & 29,403 &14,535 &290\\\hline
    France & 56,88,195 & 4,80,301 &2,59,067 &1272\\\hline
    Germany & 56,81,167 & 4,09,367 &2,42,894 &1183\\\hline
    UK & 38,26,106 & 3,46,240 &1,56,601 & 1153\\\hline
    Argentina & 6,14,872 & 59,858 &34,770 &420\\\hline
    Brazil & 10,46,409 & 1,03,687 & 69,908& 527\\\hline
    USA & 1,00,53,843 & 7,34,686 &4,26,281 &1886\\\hline
  \end{tabular}
  \caption{Table showing statistics of entities, triples, humans and occupations in each of the 13 demographics.}
  \label{tab:datset_description}
\end{table}

\section{Experiments} \label{sec:experiment}

We investigated the impact of the sensitive attribute of gender on occupations in two possible ways as mentioned earlier. First, we computed biases that exist in our KG dataset. We extended the method mentioned in \cite{bourli2020bias} to our sample dataset for measuring data bias. Next, we measured biases on two graph embedding models across the demographic dataset. For this experiment, we generated embeddings from the scratch using our giant knowledge graph. Our experimental setup closely follows the bias measurement metric as that of \cite{fisher2020measuring} but is implemented on different demographics dataset that we sampled from Wikidata. Precisely, followed by the embedding generation step based on the giant knowledge graph, we conducted bias measurement experiments separately on each demographic dataset.

\subsection{Generation of the knowledge graph embedding}
As mentioned earlier, we have employed two embedding learning models namely TransE and ComplEx in our experiment. These models are implemented using the similarity scores - L1 norm and dot product respectively. In both cases, the dimension of the graph embedding is set to 100 considering the size of the network. We train the models by fixing the negative sample size (i.e., triples) to 3 and 10 for ComplEx and TransE respectively. We restricted the number of negative samples per positive triple because of the large size of our network. The training is completed using multiclass negative log-likelihood loss as the cost function and stochastic gradient descent as the optimizer. Finally, to test the quality of the trained embeddings we evaluated them for the downstream task of link prediction using two standard metrics - Mean Reciprocal Rank (MRR) and Hits@$n$ where $n$ is set to 5, 10, and 20. In both metrics, a positive triple is ranked among 50 negative triples generated by permuting the subject/object side of the triples in the test set. For performing the evaluation, we randomly sampled a test set of $10k$ triples from the knowledge graph and computed the two metrics for both TransE and ComplEx. We performed three trials, each with a different set of randomly chosen $10k$ triples. The average result obtained for the three trials are noted in
Table~\ref{tab:test_performance}. We have ensured that the triples in every sample test set were not seen during the training phase by the embedding algorithms. We have performed our experiments\footnote{\url{https://docs.ampligraph.org/en/1.4.0/experiments.html}} with the help of well-documented library AmpliGraph\footnote{\url{https://docs.ampligraph.org/en/1.4.0/}} on our dataset.

\begin{table}
\centering
\scriptsize
  \begin{tabular}{|c|c|c|c|c|}
    \hline
    Embedding methods & MRR & Hits@5 & Hits@10 & Hits@20  \\
    \hline
    TransE & 0.683 & 0.793 & 0.841 & 0.884  \\ 
    \hline
    ComplEx & 0.774 & 0.939 & 0.978 & 0.992  \\
  \hline
\end{tabular}
\caption{Link prediction result: evaluation of pretrained embeddings for TransE and ComplEx}
\label{tab:test_performance}
\end{table}

\subsection{Data bias in Wikidata}

We studied the data bias metric for all the demographics in our dataset. The metric assigns bias score to every occupations that exists in the specific demography. Further, on the basis of bias score and threshold $t$, the occupations are categorized into three types: male-biased, female-biased and neutral. Although the set of top-ranked male and female-biased occupations vary across demographics, we found the following list of occupations ranked at top in almost all the demographics in our dataset as reported by the data bias metric.
\begin{compactitem}
    \item \textbf{male-biased occupations:} association football player, military personnel, politician, cricketer, etc.
    \item \textbf{female-biased occupations:}
    actor, writer, singer, model, etc.
    \item \textbf{neutral occupations:} 
    epidemiologist, hydrologist, social scientist, intellectual, etc. 
\end{compactitem}

\subsection{Bias measurement in KG embedding}
We applied the embedding bias metric~\cite{fisher2020measuring} (discussed briefly in the section~\ref{sec:kge_metric}) to our setting to measure the biases introduced by KG embedding methods. The metric assigns a bias score to every occupation that belongs to a specific demography and generates two individual lists of \textit{male} and \textit{female} biased occupations for a particular demography. Now, for every demography, we ranked the lists (i.e., male and female) of occupations according to their bias scores in descending order. For developing further insights, we analyzed the ranked list of occupations (male and female) based on two orthogonal axes - (i) embedding learning methods, and (ii) different demographics. 

\noindent The goal of our experiments is to answer the following questions -
\begin{compactitem}
    \item How do the two embedding learning methods - TransE and ComplEx rank biased occupations for a given demography? Are two rankings generated by two models similar or very different from one another?
    \item How do the rankings of gender-biased occupations vary across demographics? Do we observe any geographical differences among these ranked lists? 
\end{compactitem}
The similarity between different rankings is compared using the Jaccard similarity metric. This is done separately for each demography.

\section{Key results} \label{sec:result}

In this section, we present the answers to the questions that we posed above.

\subsection{Data bias vs algorithmic bias} \label{sec:data vs algorithm}

Here we attempt to corroborate the first hypothesis - the biases present in the base data are altered by the embedding learning algorithms. For this purpose, we compute a rank list of male and female-biased occupations based on the (i) data bias metric and the (ii) bias metric for KG embedding. For the top $K$ occupations based on the data bias metric, we attempt to find their corresponding ranks in the second list of ranking based on the embedding bias metric and compute the rank deviation. Let us assume, a occupations $p$ is ranked at $r_{1}$ and $r_{2}$ in the two rankings produced by the data bias and the embedding bias metrics respectively. We took the inverse of the ranks and calculated rank deviation as follows - $(1/r_{1}) - (1/r_{2})$ \footnote{Simple $r_{1} - r_{2}$ wouldn't differentiate between pairs \{$r_{1} = 1$, $r_{2}=2$\}, and \{$r_{1}=100$, $r_2=101$\} i.e reordering amongst the top ranked vs low ranked occupations.}. Finally, the overall rank deviation of the two rank lists is computed by averaging the individual rank deviation of the occupations that are ranked at top $k$ positions based on the data bias metric. A positive (negative) value of rank deviation indicates that occupations are ranked at a lower (higher) position in the second rank list (i.e., bias metric for KG embedding) relative to the first one (i.e., data bias metric). The higher the deviation (on either positive or negative sides), the stronger is the evidence that the inherent data bias is altered by the embedding learning algorithms.

\noindent For a given demography $D$, the rank deviation is computed for the following pairs of the ranked list of occupations ranked at $K$ based on data bias metric.
\begin{compactitem}
    \item male-biased occupations by data bias metric \textit{vs.} male-biased occupations by TransE's ranking using embedding bias metric.
    \item female-biased occupations by data bias metric \textit{vs.} female-biased occupations by TransE's ranking using embedding bias metric.
    \item male-biased occupations by data bias metric \textit{vs.} male-biased occupations by ComplEx's ranking using embedding bias metric.
    \item female-biased occupations by data bias metric \textit{vs.} female-biased occupations by ComplEx's ranking using embedding bias metric.
\end{compactitem}

The rank deviations for both male and female occupations across the demographics are noted in Table~\ref{tab:rank_deviation}. The rank deviation values show that  
the two rankings generated by the data bias metric and the embedding bias metric are characteristically different for $K = 20$. Here the positive value of rank deviation indicates that the embedding learning methods ranked the top occupations according to the data bias metric at relatively lower positions. Further, almost all the demographics exhibited similar rank deviation. Thus we establish that the inherent data bias present in the dataset is indeed altered by the embedding learning algorithms. The results further indicate that if one discounts the inherent data bias, the embedding learning algorithms themselves introduce biases that can potentially affect the downstream NLP applications that rely on KG embeddings. Thus the pertinent question next is whether the biases introduced by the embedding learning algorithms are equivalent or different. This is what we attempt to answer in the next section.
\begin{table}[h]
\scriptsize
    \centering
    \begin{tabular}{|p{2cm}||c|c|c|c|}
    \hline
    \multirow{2}{*}{Demography} & \multicolumn{2}{c|}{TransE} & \multicolumn{2}{c|}{ComplEx} \\
    \cline{2-5}
    & \multicolumn{1}{c|}{Male} & \multicolumn{1}{c|}{Female} & \multicolumn{1}{c|}{Male} & \multicolumn{1}{c|}{Female} \\
    \cline{2-5}
    \hline
    Arabia & 0.15 & 0.13 & 0.11 & 0.12  \\
    \hline
    India & 0.17 &  0.13 &  0.17 &  0.10 \\
    \hline
    Japan & 0.16 & 0.11 & 0.16 & 0.10 \\
    \hline
    Russia & 0.16 & 0.12 & 0.12 & 0.15 \\
    \hline
    Australia & 0.16 & 0.15 & 0.11 & 0.14  \\
    \hline
    Kenya & 0.07 & 0.10 & 0.12 & 0.08 \\
    \hline
    South Africa & 0.10 & 0.14 & 0.14 & 0.16 \\
    \hline
    France & 0.15 & 0.13 & 0.17 & 0.10 \\
    \hline
    Germany & 0.18 & 0.09 & 0.11 & 0.17 \\
    \hline
    UK & 0.16 & 0.09 & 0.14 & 0.16 \\
    \hline
    Argentina & 0.13 & 0.13 & 0.12 & 0.12\\
    \hline
    Brazil & 0.14 & 0.13 & 0.10 & 0.10 \\
    \hline
    USA & 0.18 & 0.16 & 0.16 & 0.14 \\
    \hline
    \end{tabular}
    \caption{Rank deviation (as explained in section~\ref{sec:data vs algorithm}) of the two ranked lists generated by data bias metric and KG embedding bias metric truncated at the top ($K$ = 20) ranked occupations.}
    \label{tab:rank_deviation}
\end{table}

\subsection{Comparison of two embedding methods: TransE with ComplEx}
\label{sec:comparison_transe_complex}

 In this section, we attempt to identify whether the biases introduced due to the embedding learning algorithms are the same or different across these algorithms. If they are indeed different then the downstream applications would differ based on the type of embedding algorithm used. For the two different categories of biased occupations (i.e., male-biased and female-biased), we computed the similarity between the two ranked lists of occupations generated by the embedding methods - TransE and ComplEx. The comparison has been iterated over all the 13 demographics in our dataset and the result is shown in Table~\ref{tab:sim_embedding}. Overall we found very little overlap between the two rankings of biased occupations obtained by the two embedding methods. Because of the difference in scoring functions, the embedding algorithms learn the underlying graph structure differently and perhaps influence the dissimilar ranking of our sample set of occupations. We computed this overlap for varying values of top $K$ occupations in the rank list. For all values of $K$ (20, 50, and 80) we observed a similar trend of minimal overlap with very low values of Jaccard similarity for both male and female-biased occupations. On a closer inspection into the list of biased occupations ranked by TransE and ComplEx, we found that TransE pulls up very generic occupations at the top of the list. In contrast, demography-specific occupations are ranked at higher positions in the list pulled up by the ComplEx embedding. For example, a list of occupations ranked at the top 50 by the two independent embedding methods are -
\begin{compactitem}
    \item \textbf{male biased occupations ranked by TransE:} sportsman, specialist, climber, referee, leader etc.
    \item \textbf{male biased occupations ranked by ComplEx:} adventurer, agent, charlatan, clergy, intelligence agent etc. 
    \item \textbf{female biased occupations ranked by TransE:} activist, expert, discussion moderator, occupationsal, erudite etc. 
    \item \textbf{female biased occupations ranked by ComplEx:} feminist, traveler, faculty, home keeper, aviation employees etc.
\end{compactitem}
While in general, all overlaps are low between the biased rank lists generated by the two embeddings, we also observed that a few of the demographics such as Germany, France, UK, USA have exceptionally low overlap (e.g., highlighted in red in Table ~\ref{tab:sim_embedding}). Certain other demographics like Arabia, Kenya, South Africa have a relatively higher overlap between the biased rank lists generated by the two embeddings (highlighted in green in Table~\ref{tab:sim_embedding}) in our dataset. The analysis revealed that the two embedding methods follow a different portfolio of biases in ranking biased occupations. In a nutshell, the sensitive attribute gender has a varying impact on listing biased occupations when one employs different graph embedding models to embed the underlying knowledge graph.

In order to investigate the generalizability of the results we computed the results for a third embedding learning algorithm -- DistMult~\cite{yang2014embedding} (see Appendix) and found that our observations about the low overlap across different learning algorithms hold true even though ComplEx and DistMult are from the same family.  
\begin{table}
\centering
\scriptsize
  \begin{tabular}{|p{1.3cm}||c|c|c|c|c|c|}
    \hline
    \multirow{2}{*}{Demography} & \multicolumn{2}{c|}{K=20} & \multicolumn{2}{c|}{K=50} & \multicolumn{2}{c|}{K=80} \\
    \cline{2-7}
    & Male & Female & Male & Female & Male & Female \\
    \hline
    \rowcolor{green!30}
    Arabia & 0.11 & 0.14 & 0.16  & 0.22 & 0.34 & 0.31 \\
    \hline
    India & 0.03 & 0.08 & 0.08 & 0.12 & 0.09 & 0.16 \\
    \hline
    Japan & 0.05 & 0.03 & 0.03 & 0.06 & 0.05 & 0.13 \\
    \hline
    Russia & 0.03 & 0.03 & 0.01 & 0.08 & 0.03& 0.14\\
    \hline
    Australia & 0.03 & 0.0 & 0.03 & 0.03 & 0.05 & 0.11\\
    \hline
    \rowcolor{green!30}
    Kenya & 0.05 & 0.18 & 0.3 & 0.3 & 0.7 & 0.63\\
    \hline
    \rowcolor{green!30}
    South Africa & 0.05 & 0.05 & 0.1 & 0.18 & 0.11 & 0.21\\
    \hline
     \rowcolor{red!30}
    France & 0.00 & 0.03 & 0.02 & 0.04 & 0.02 & 0.08\\
    \hline
    \rowcolor{red!30}
    Germany & 0.00 & 0.00 & 0.01 & 0.05 & 0.01 & 0.07\\
    \hline
    \rowcolor{red!30}
    UK & 0.00 & 0.00 & 0.01 & 0.05 & 0.03 & 0.08\\
    \hline
    Argentina & 0.05 & 0.08 & 0.04  & 0.14 & 0.07 & 0.17\\
    \hline
    Brazil & 0.05 & 0.00 &0.06 & 0.12 & 0.08 & 0.14\\
    \hline
    \rowcolor{red!30}
    USA & 0.00 & 0.00 & 0.01  & 0.03 & 0.01 & 0.05\\
    \hline
    \end{tabular}
    \caption{Jaccard similarity between the lists of biased occupations (male and female both) ranked by two different embedding techniques - \textbf{TransE} and \textbf{ComplEx}. The results are obtained for the lists ranked at the top $K$ (20, 50 and 80) biased occupations for all the demographics. The rows highlighted in \textcolor{red}{red} indicate very less similarity in the ranking of biased occupations obtained from the two embedding methods for all values of $K$. In contrast,  the \textcolor{green}{green} highlighted rows for all values of $K$ denote relatively slightly higher similarity of biased occupations ranked based on the two embedding methods.}
  \label{tab:sim_embedding}
\end{table}

\subsection{Comparison of different demographics}

Similar to the comparison of biased occupations based on embedding methods, we compared the rankings across the 13 demographics in our dataset. First, we generated the lists of male and female-biased occupations for each demography using TransE and ComplEx. This leads us to 4 different ranked lists for every demography depending on the pairwise combination of the sensitive attributes - male/female and two embedding methods. Formally, for a demography $D$, the list of biased occupations $L_{1}$ (TransE-male), $L_{2}$ (TransE-female), $L_{3}$ (ComplEx-male), $L_{4}$ (ComplEx-female) are assumed to be ranked based on their bias scores and top 20 occupations are selected from each of the list for further analysis. For every demography, we performed the same exercise leading us to 4 different lists for each of them. Now, for demography $D$, we compared the 4 lists namely $L_{1}$, $L_{2}$, $L_{3}$ and $L_{4}$ ranked at the top 20 with the corresponding lists of remaining 12 demographics. This produces a similarity index for each pair of the lists thus generating 12 similarity values. Finally, we average these values and compute the mean and standard deviation for demography $D$. The result is tabulated in Table~\ref{tab:avg_sim_demography}.

\noindent We observed that average pairwise similarity across all the demographics for the ComplEx method is much lower than TransE. For example, the USA shows average an similarity of 0.32 and 0.15 with the remaining demographics for male-biased occupations when the embeddings are generated using TransE and ComplEx respectively. This phenomenon also holds for female-biased occupations. Further, we repeated the same procedure for the top occupations ranked at 50 and 80 and it produces a very similar trend. Thus, we believe is a corollary of the fact that ComplEx pulls up very demographic-specific occupations (which are therefore different across demographics resulting in low overlap) while TransE pulls up generic occupations common across the different demographics (resulting in a higher overlap).

\begin{table}[h]
\centering
\scriptsize
    \begin{tabular}{|c|p{0.4cm}|p{0.4cm}|p{0.4cm}|p{0.4cm}|p{0.4cm}|p{0.4cm}|p{0.4cm}|p{0.4cm}|}
    \hline
    \multirow{3}{*}{Demography} & \multicolumn{4}{c|}{TransE} & \multicolumn{4}{c|}{ComplEx} \\
    \cline{2-9}
    & \multicolumn{2}{c|}{Male} & \multicolumn{2}{c|}{Female} & \multicolumn{2}{c|}{Male} & \multicolumn{2}{c|}{Female} \\
    \cline{2-9}
    & mean & std. dev & mean & std. dev & mean & std. dev & mean & std. dev\\
    \hline
    Arabia & 0.10 & 0.07 & 0.38  & 0.08 & 0.08 & 0.05 & 0.12 & 0.05   \\
    \hline
    India & 0.17 & 0.04 & 0.50 & 0.13 & 0.19 & 0.09 & 0.26 & 0.14 \\
    \hline
    Japan & 0.30 &0.16 &0.35 & 0.07& 0.17& 0.07 & 0.21 & 0.15 \\
    \hline
    Russia & 0.31 & 0.19 & 0.51 & 0.15 & 0.17 & 0.08 & 0.23 & 0.16 \\
    \hline
    Australia & 0.32 & 0.15 & 0.56 & 0.17 & 0.19 & 0.08 & 0.26 & 0.13\\
    \hline
    Kenya & 0.06 & 0.07 & 0.23 &  0.05 & 0.05 & 0.03 & 0.06 & 0.03\\
    \hline
    South Africa & 0.18 & 0.06 & 0.5 &0.13 & 0.10 & 0.04 & 0.27 & 0.15\\
    \hline
    France & 0.24 & 0.14 & 0.53 & 0.18 & 0.21 & 0.13 &0.21 & 0.12\\
    \hline
    Germany &0.30 & 0.19 &0.35 & 0.07 & 0.19 & 0.13 & 0.25 & 0.12\\
    \hline
    UK & 0.32 & 0.18 & 0.54 & 0.18 & 0.20 & 0.12 & 0.27 & 0.15\\
    \hline
    Argentina & 0.26 & 0.13 & 0.48 & 0.15 & 0.15 & 0.09 & 0.26 & 0.16\\
    \hline
    Brazil & 0.25 & 0.13 & 0.47 & 0.12 & 0.21 & 0.08 & 0.29 & 0.14\\
    \hline
    USA & 0.32 & 0.21 & 0.32 & 0.10 & 0.15 & 0.10 & 0.23 & 0.15\\
    \hline
    \end{tabular}
    \caption{Average similarity (mean, std. deviation) across demographics for biased occupations (male and female individually) ranked at top 20. The similarity has been computed for both the embedding methods TransE and ComplEx.}
    \label{tab:avg_sim_demography}
\end{table}

\begin{figure*} [h]
     \centering
     \begin{subfigure}[b]{125pt}
         \centering
         \includegraphics[width=\textwidth]{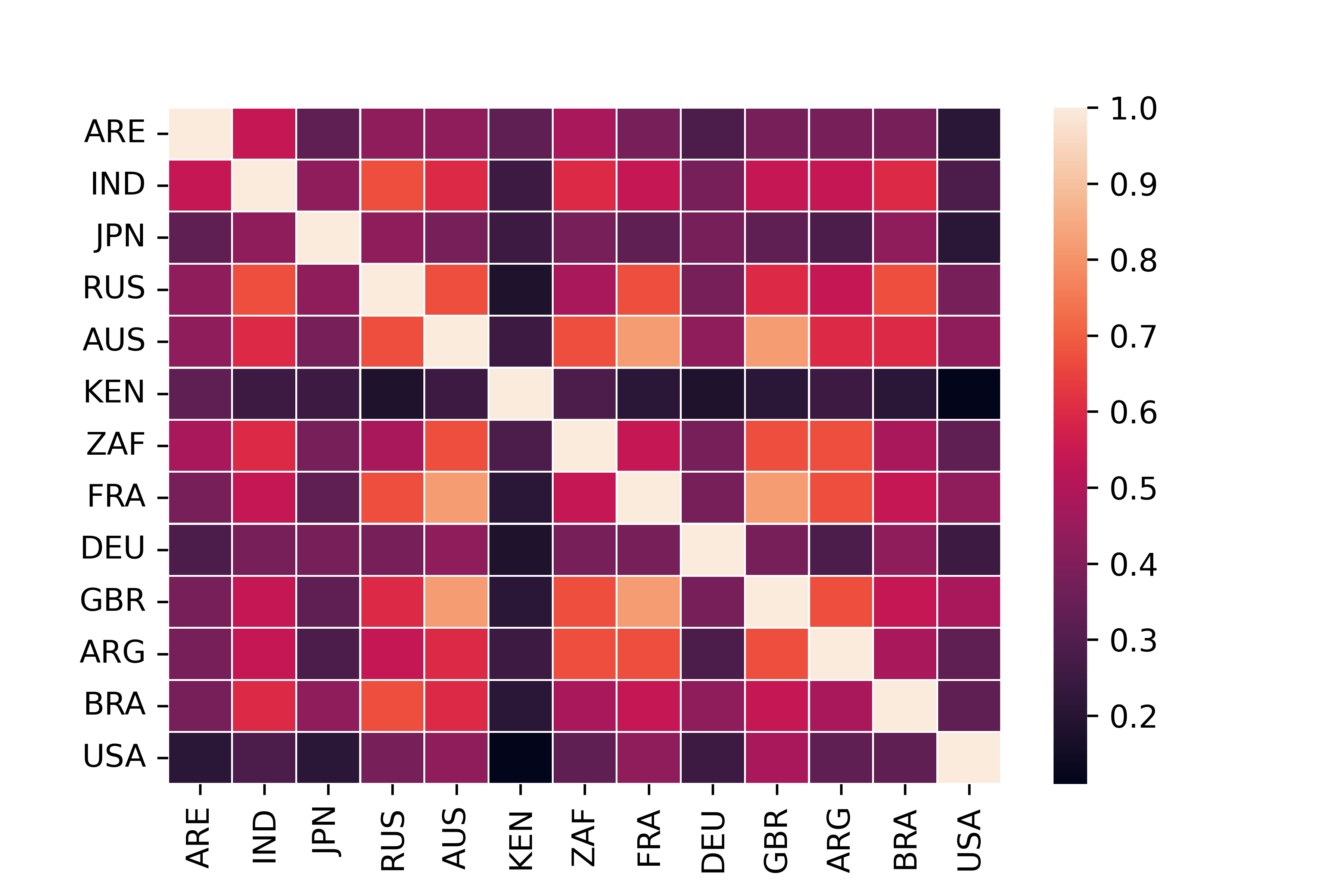}
         \caption{$TransE: Female$}
         \label{fig:female_transE}
     \end{subfigure}
     \hfill
     \begin{subfigure}[b]{125pt}
         \centering
         \includegraphics[width=\textwidth]{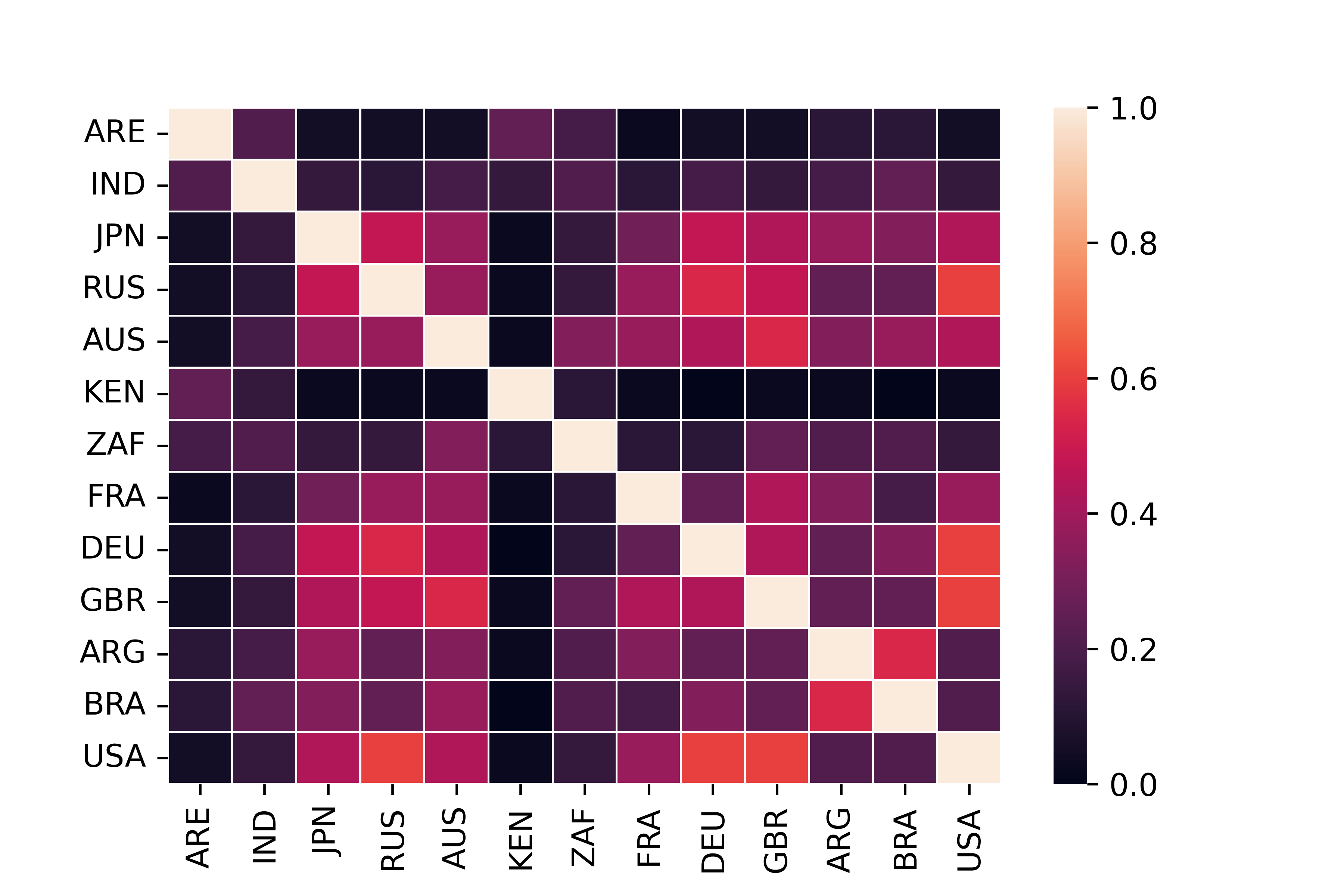}
         \caption{$TransE: Male$}
         \label{fig:male_trasnE}
     \end{subfigure}
     \hfill
     \begin{subfigure}[b]{125pt}
         \centering
         \includegraphics[width=\textwidth]{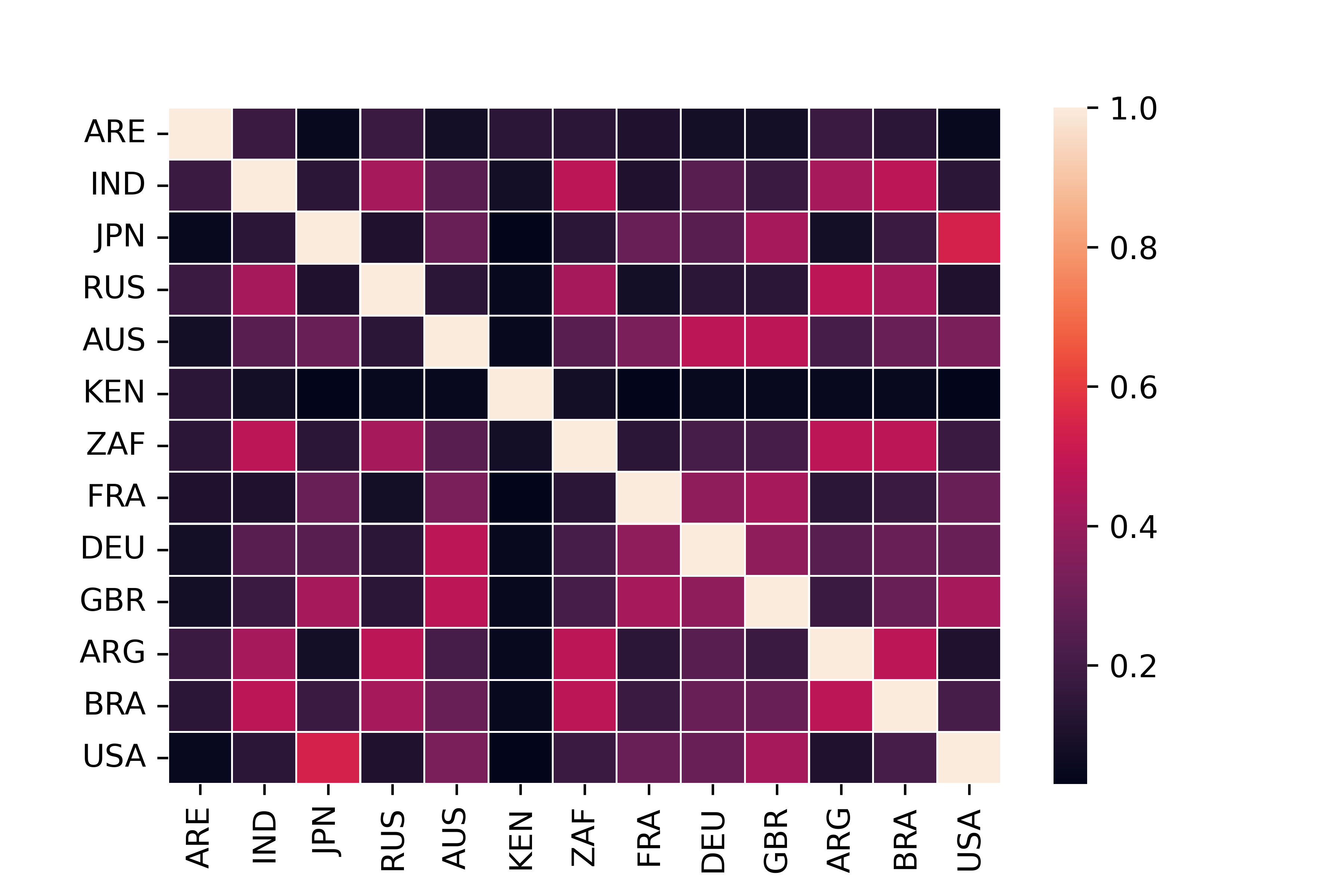}
         \caption{$ComplEx: Female$}
         \label{fig:female_complEx}
     \end{subfigure}
     \hfill
     \begin{subfigure}[b]{125pt}
         \centering
         \includegraphics[width=\textwidth]{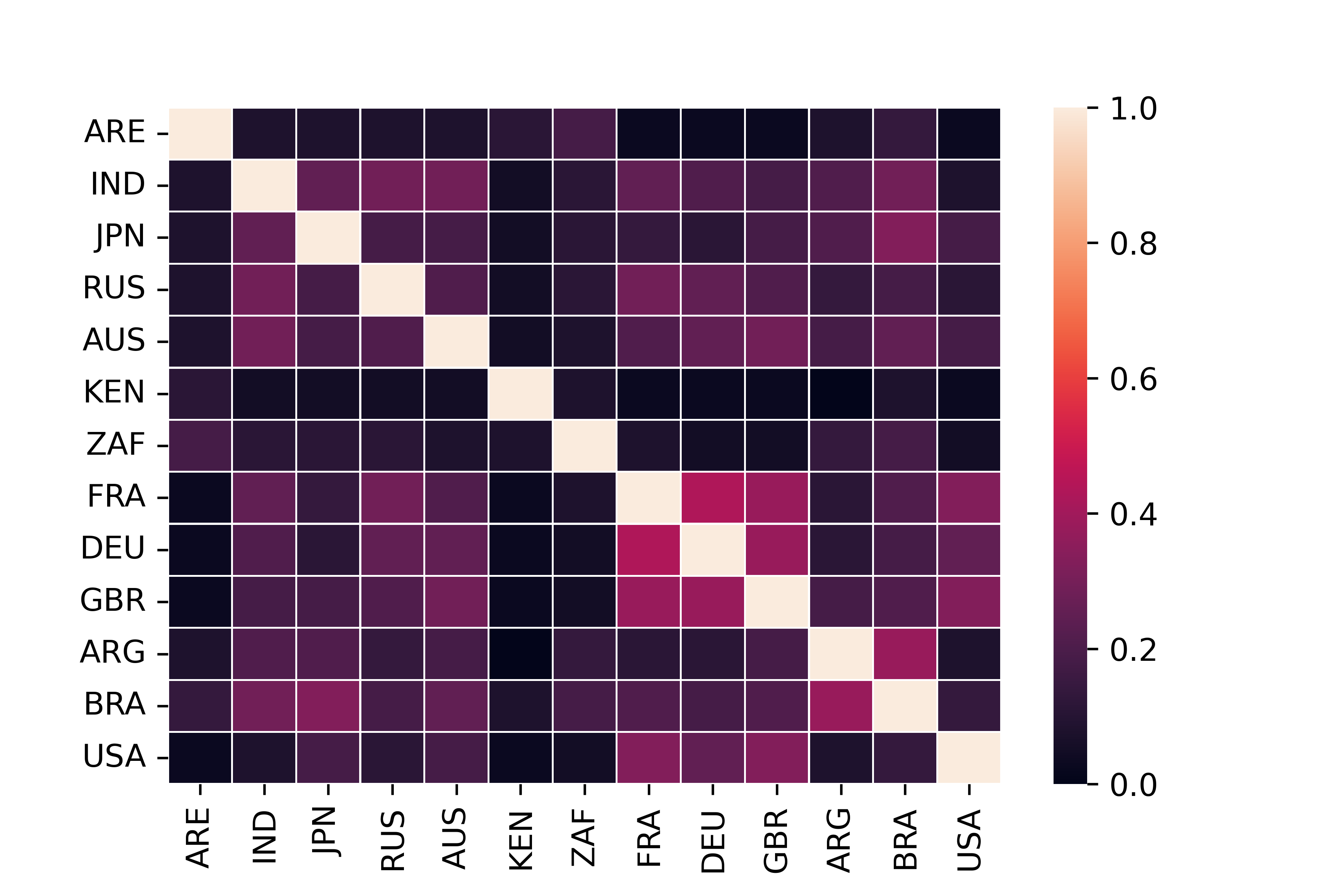}
         \caption{$ComplEx: Male$}
         \label{fig:male_ComplEx}
     \end{subfigure}
     \caption{Heatmaps showing similarity between different demographics for the male and female-biased occupations ranked at $K=20$. The heatmaps in the top and bottom rows are generated for TransE and ComplEx respectively. The abbreviations used for the names of the countries are as per the ISO 3166 standard.}
        \label{fig:comp_demographics}
\end{figure*}

\noindent\textit{\textbf{Average similarity between demographics:}} The heatmaps in Figure~\ref{fig:comp_demographics} present an illustration of average similarity between every pairs of demographics for the four combinations: TransE-male, TransE-female, ComplEx-male, ComplEx-female. The figures make
it visually evident that TransE exhibits more similarity for a significant number of pairwise demographics compared to ComplEx. Further, the figures depict that female-biased occupations are more similar across demographics than the male-biased occupations. This supports our insights presented in Table~\ref{tab:avg_sim_demography} (please refer to the columns named \textit{female} for both TransE and ComplEx). 

\noindent\textit{\textbf{Most similar demographics:}} In addition, we carried out further analysis to find for a given demography what are the topmost similar demographics (top three) across each of the combinations (i.e., TransE-male, TransE-female, ComplEx-male, ComplEx-female) presented in the heatmaps (please see Figure~\ref{fig:comp_demographics}). The results are noted in Table~\ref{tab:similar_demo}. As has been observed so far, this list of the top three closest demographics varies widely based on the embedding method for each and every demography. Further for a particular embedding method and particular demography, the top three closest demographics as per male-biased and female-biased occupations are also different. One interesting point that we note is that the UK features in the list of top three for all four combinations in the case of at least three different demographics -- USA, Australia, and France. Manual inspection shows that the occupations that connect UK to these demographics are rugby union player (male), rabbi (male), scientific illustrator (female), and suffragist (female) which possibly corresponds to the shared cultural heritage and economic parity of these demographics with the UK. In Table~\ref{tab:pair_demographics} we note the most similar pair of demographics and manually inspect whether this similarity can be attributed to either socio-cultural or geographical closeness. TransE pulls up pairs like (UK, USA), (Australia, UK), (India, South Africa) as similar. These pairs are geographically quite apart sometimes even appearing in opposite hemispheres. The reason for their similarity can be primarily attributed to their close socio-economic and socio-cultural complexion. Since TransE pulls up biased occupations that are generic, this socio-economic and socio-cultural closeness gets naturally manifested. All the demography pairs that we manually find to be close in this dimension are marked in red in Table~\ref{tab:pair_demographics}. On the other hand, certain pairs that are geographically close are pulled up as similar by ComplEx. Closer geographies should have similar types of nuances in their list of most biased occupations which is what is extracted by ComplEx and hence the similarity (observed more across the male axis compared to the female axis). We mark all these cases in blue in Table~\ref{tab:pair_demographics}.

\section{Discussion} \label{sec:discussion}
The primary objective of this work has been to bring forth the fine-grained details of biases manifesting from the disparity of data representation across the different demographics as well as the non-neutral learned embeddings from knowledge graphs. In this section, we discuss a few other interesting insights that we obtained from our analysis. 

\subsection{Unique occupations per demography}

In an ideal scenario, occupations should be free from sociocultural and geographic delineations. We revisit the top listed occupations (ranked at 100) for each demography and tried to find unique occupations that signify demography-specific characteristics. 
A closer inspection of this list shows that the occupations pulled up by ComplEx embedding resemble nuanced examples of demography-specific characteristics. For example, the occupations \textit{sitarist} for India, \textit{Islamic jurist} for Arabia, \textit{bhikkhu} for Japan symbolize the cultural heritage of a specific geographic area. In contrast, the occupations selected by TransE mostly represent different sports-related occupations for male and media-related occupations for females. Moreover, some controversial occupations such as \textit{playboy playmate} (USA), \textit{pornographic film director} (Russia), \textit{witch} (Kenya), and \textit{cunning folk} (UK) emerge as female-dominated occupations and \textit{terrorist} (Arabia), \textit{militant} (Arabia), \textit{holocaust denier} (Germany) and \textit{anarchist} (Argentina) emerge as male-dominated occupations. These need to be  judiciously audited by the Wikipedia community at large in order to promote a safer cyberspace. 

\begin{table*}
\centering
\scriptsize
    \begin{tabular}{|p{2cm}||p{2.5cm}|p{2.5cm}|p{2.5cm}|p{2.5cm}|}
    \hline
    \multirow{2}{*}{Demography} & \multicolumn{2}{c|}{TransE} & \multicolumn{2}{c|}{ComplEx} \\
    \cline{2-5}
    & Male & Female & Male & Female \\
    \hline
    Arabia & Kenya, India, South Africa & India, South Africa, Russia & South Africa, Brazil, Kenya & India, Russia, Argentina\\
    \hline
    India & Brazil, Arabia, South Africa & Russia, South Africa, Brazil & Russia, Brazil, Japan & South Africa, Brazil, Russia\\
    \hline
    Japan & Russia, Germany, UK & India, Russia, Brazil & Brazil, India, Argentina & USA, UK, Australia \\
    \hline
    Russia & USA, Germany, Japan & India, Australia, France & India, France, Germany & Argentina, India, South Africa\\
    \hline
    Australia & UK, Germany, Japan & France, UK, Russia & UK, India, Germany & Germany, UK, France\\
    \hline
    Kenya & Arabia, India, South Africa & Arabia, South Africa, India & Arabia, South Africa, Brazil & Arabia, India, South Africa \\
    \hline
    South Africa & Australia, Brazil, Argentina & Australia, UK, Argentina & Arabia, Brazil, Argentina & India, Argentina, Brazil\\
    \hline
    France & UK, USA, Russia & Australia, UK, Russia & Germany, UK, USA & UK, Germany, Australia\\
    \hline
    Germany & USA, Russia, Japan & Australia, India, Japan & France, UK, USA & Australia, France, UK\\
    \hline
    UK & USA, Australia, Russia & Australia,France,South Africa & Germany, USA, Australia & Australia, Japan, USA \\
    \hline
    Argentina & Brazil, Japan, Australia & South Africa, France, UK & Brazil, India, Japan & Russia, South Africa, Brazil\\
    \hline
    Brazil & Argentina, Australia, Japan & Russia, India, Australia & Argentina, Japan, India &  India, South Africa, Argentina\\
    \hline
    USA & UK, Germany, Russia & UK, Australia, France & UK, France, Germany & Japan, UK, Australia\\
    \hline
    \end{tabular}
    \caption{Table showing demographics that exhibit similarity in ranking biased occupations (ordered in decreasing value of similarity in each cell).}
  \label{tab:similar_demo}
\end{table*}

\begin{table}
\scriptsize
  \begin{tabular}{|p{1.5cm}|p{3cm}|p{3cm}|}
    \hline
    Demography & Male & Female \\
    \hline
    TransE & \textcolor{red}{(UK, USA)}, \textcolor{red}{(Australia,UK)}, \textcolor{red}{(Russia, Japan)}, \textcolor{red}{(Russia, UK)} & (France, UK), (India, Russia), \textcolor{red}{(India, South Africa)}, \textcolor{red}{(Australia, UK)}\\
    \hline
    ComplEx & \textcolor{blue}{(France, Germany)}, \textcolor{blue}{(France, UK)}, \textcolor{blue}{(Germany, UK)}, \textcolor{blue}{(Argentina, Brazil)} & (Australia, Germany), (Australia, UK), (India, South Africa), \textcolor{blue}{(Argentina, Brazil)}  \\
  \hline
\end{tabular}
\caption{Table showing most similar pairs of demographics for the combinations- TransE-male, TransE-female, ComplEx-male, ComplEx-female.}
  \label{tab:pair_demographics}
\end{table}

\subsection{Female biased occupations are less diverse}

\begin{figure} [h]
    \centering
    \begin{subfigure} [b]{0.23\textwidth}
        \centering
        \includegraphics[width=\textwidth]{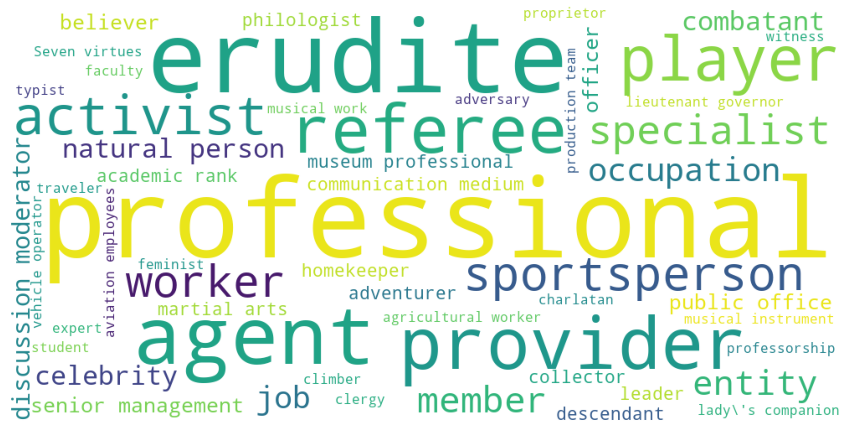}
        \caption{Word cloud: Female}
        \label{fig:words_female}
    \end{subfigure}
    \hfill
    \begin{subfigure} [b]{0.23\textwidth}
        \centering
        \includegraphics[width=\textwidth]{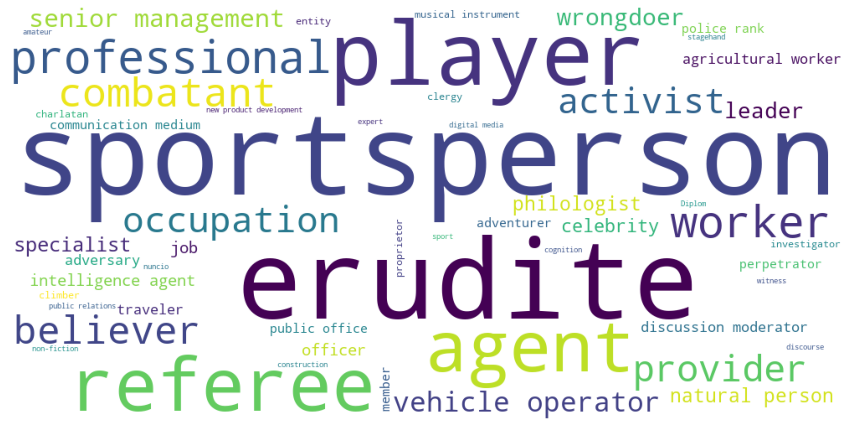}
        \caption{Word cloud: Male}
        \label{fig:words_male}
    \end{subfigure}
    \caption{Word clouds showing distinct (a) female-biased and (b) male-biased occupations.}
    \label{fig:words_cloud}
\end{figure}

Although the Wikimedia foundation provides an open-source gender-neutral editing environment to all the digitally literate people around the globe, it is now well-known that there exists a substantial gender gap in many Wikimedia projects in terms of the number of editors and editing ractices~\cite{konieczny2018gender,collier2012conflict,wagner2015s}. Further, researchers~\cite{sun2021men} discovered implicit gender biases in describing human biographies. Similar to other Wikimedia projects, gender inequality and under representation of content about female compared to male exists in Wikidata as well in varying degrees. In fact, a recent study~\cite{zhang2021quantifying} found that only 22\% of Wikidata items that represent people are about women. Likewise, our detailed analysis has identified several instances in which male-biased occupations exhibit a diverse range of occupations (see Figure~\ref{fig:words_male}) across different demographics. In contrast, female-biased occupations are confined to similar types of occupations mostly, for example, occupationsal, activist, erudite, etc. (see Figure~\ref{fig:words_female}). Further, the biased occupations in each type (male and female) are represented in such a way that it reflects the so-called gender discrimination that persists in our society. For example, females are categorized into a typical generic set of occupations, such as activist, specialist, home keeper, etc. whereas males are subjected to the role specific occupations, e.g., leader, vehicle operator, believer etc (as depicted in the Figure~\ref{fig:words_cloud}). To quantify this gender disparity with regard to the representation of occupations across demographics, we measured the diversity of individual groups of male and female-biased occupations as follows.

\noindent First we collect the list of the top 50 biased occupations from each of the demographics and combined them into a set of the distinct vocabulary of occupations separately for the male and the female categories. We calculate the probability ($p_{prof}$) of the occurrence of each of the biased occupations ($prof$) in the 13 demographics of our dataset. Now, we calculate the Shannon entropy for each of the occupations ($-p_{prof}\log{p_{prof}}$) and finally sum over all the occupations ($-\sum_{prof} p_{prof}\log{p_{prof}}$). This process is iterated on both male and female categories of biased occupations for the two embedding methods separately. A higher entropy value would indicate a larger diversity of occupations over the demographics. The obtained entropy values are shown in Table~\ref{tab:entropy}. The entropy measure clearly illustrates that the male-biased occupations exhibit a larger variety than female-biased ones for both the embedding generation methods. Perhaps Wikidata editors tend to add  certain types of occupations that substantially favour masculinity over feminism which in turn reflects the real world gender bias. 

\begin{table}[h]
  \begin{center} 
  \begin{tabular}{|c|c|c|}
    \hline
    Gender & TransE & ComplEx \\
    \hline
    Male & 55.69 & 68.98 \\
    \hline
    Female & 41.96 & 50.32 \\
  \hline
\end{tabular}
  \end{center}
\caption{Table showing entropy based diversity for representing distinct male and female-biased occupations.}
  \label{tab:entropy}
\end{table}

\section{Conclusion} \label{sec:conclusion}

In this paper, we have performed a large-scale audit to investigate the presence of data bias in the knowledge graph (i.e, Wikidata) by leveraging diverse distribution of dataset (i.e, sub-sample of 13 demographics) and algorithm bias in KG embedding by applying two distinct embedding models-- TransE and ComplEx. We showed that diversity in the dataset and embedding methods with regard to specific sensitive features (e.g., the impact of gender on occupation) can result in stark differences in the ranking of biased occupations. As a future work, we will extend the proposed approach to multi-valued sensitive attributes. This will help us to build a system that will demonstrate the fine-grained distribution of societal biases across the globe and take suitable measures to mitigate them.

\bibliographystyle{ACM-Reference-Format}
\bibliography{ref}

\appendix
\section{Appendix}
As an extension of the quantitative study of bias exploration in Wikidata, we conducted experiments with another embedding learning algorithm, DistMult~\cite{yang2014embedding} which belongs to the tensor decomposition model category~\cite{rossi2021knowledge}. The embedding is learned on the same large network consisting of thirteen demographics and the biases exhibited by the algorithm are reported in this section. Similar to the previous experiment of evaluation of learned embedding for TransE and ComplEx, we computed MRR and Hits@n (i.e, $n$ is set to 5, 10 and 20) for ranking a positive triple among 50 negative triples and the obtained metrics-- MRR, Hits@5, Hits@10, and Hits@20 are 0.23, 0.49, 0.73, 0.84 respectively.

\subsection{Data bias vs algorithm bias}
As described in the section~\ref{sec:data vs algorithm}, we find out the rank deviations of two rankings of biased occupations, truncated at the top 20 ranks, generated by the data bias metric and the embedding bias metric for the embedding learning algorithm DistMult. The result is tabulated in Table~\ref{tab:DistMult_data_algo}. It follows a similar trend as shown in Table~\ref{tab:rank_deviation}. Likewise TransE and ComplEx, the positive value of rank deviation for each demography signifies that DistMult too admits algorithmic bias which is different from the data bias.

\begin{table}
    \centering
    \scalebox{0.8}{
    \begin{tabular}{|p{2cm}|c| c|}
    \hline
         Demography & Male & Female  \\
        \hline
         Arabia & 0.16 & 0.09\\
         \hline
         India & 0.17 & 0.08\\
         \hline
         Japan & 0.16 & 0.16 \\
         \hline
         Russia & 0.16 & 0.10 \\
         \hline
         Australia & 0.18 & 0.11 \\
         \hline
         Kenya & 0.15 & 0.13 \\
         \hline
         South Africa & 0.17 & 0.09 \\
         \hline
         France & 0.15 & 0.11 \\
         \hline
         Germany & 0.16 & 0.08 \\
         \hline
         UK & 0.18 & 0.09 \\
         \hline
         Argentina & 0.17 & 0.11 \\
         \hline
         Brazil & 0.17 & 0.11 \\
         \hline
         USA & 0.18 & 0.09 \\
         \hline
    \end{tabular}
    }
    \caption{Rank deviation of the two ranked lists generated by data bias metric and KG embedding bias metric ranked for the embedding learning algorithm \textbf{DistMult} truncated at top ($K=20$) occupations.}
    \label{tab:DistMult_data_algo}
\end{table}

\subsection{Comparison with embedding learning algorithms - TransE and ComplEx}
We attempted to identify the similarity of DistMult with two prime embedding learning algorithms in our work, TransE, and ComplEx. The comparison is done for each of the demographics for both male and female-biased occupations ranked at top 80 positions in the list as generated by the algorithms. As mentioned in the section~\ref{sec:comparison_transe_complex}, we computed the overlap in terms of Jaccard similarity and the corresponding values for each of the algorithms -- TransE and Complex are reported in the Table~\ref{tab:jaccard_distmult}.
The result shows that the embedding algorithm DistMult has a very low overlap with the other two embedding algorithms. We have observed a similar trend for different values of $K$ (i.e., 20, 50) which once again supports our hypothesis of variations in the bias profiles exhibited by different embedding learning algorithms. Even though DistMult is from the same family as ComplEx, their bias profiles vary a lot, especially in the case of males. Further, a closer investigation into the list of biased occupations as ranked by DistMult reveals a combination of occupations categorized as biased -- demography-specific as well as generic in characteristics. A list of such gender-biased occupations ranked by DistMult is mentioned below.
\begin{itemize}
    \item \textbf{male-biased occupations}: Islamic jurist, peace activist, comedian, cricketer, sculptor, sertanista, etc.
    \item \textbf{female-biased occupations}: art critic, receptionist,  journalist, musician, acrobatic gymnast, canoeist, etc.   
\end{itemize}

\begin{table}[h]
    \centering
    \scalebox{0.8}{
    \begin{tabular}{|p{1.8cm}||p{0.8cm}|p{0.8cm}|p{0.8cm}|p{0.8cm}|}
    \hline
    \multirow{2}{*}{Demography} & \multicolumn{2}{c|}{TransE} & \multicolumn{2}{c|}{ComplEx}\\
    \cline{2-5}
    & Male & Female & Male & Female \\
    \hline
    Arabia & 0.21 & 0.3 & 0.28 & 0.34\\
    \hline
    India & 0.05 & 0.19 & 0.11 & 0.17\\
    \hline
    Japan & 0.02 & 0.15 & 0.03 & 0.13 \\
    \hline
    Russia & 0.04 & 0.17 & 0.03 & 0.11\\
    \hline
    Australia & 0.01 & 0.16 & 0.03 & 0.09 \\
    \hline
    Kenya & 0.67 & 0.65 & 0.72 & 0.70\\
    \hline
    South Africa & 0.12 & 0.27 & 0.16 & 0.17\\
    \hline
    France & 0.01 & 0.13 & 0.01 & 0.04 \\
    \hline
    Germany & 0.03 & 0.13 & 0.01 & 0.08  \\
    \hline
    UK & 0.01 & 0.11 & 0.00 & 0.10\\
    \hline
    Argentina & 0.06 & 0.19 & 0.09 & 0.12\\
    \hline
    Brazil & 0.04 & 0.22 & 0.05 & 0.12\\
    \hline
    USA & 0.00 & 0.11 & 0.01 & 0.05 \\
    \hline
    \end{tabular}
    }
    \caption{Similarity of \textbf{DistMult} with the lists of biased occupations (male and female both) ranked at top $K$ (e.g.,80) by the embedding technique - \textbf{TransE} and \textbf{ComplEx}}. 
    \label{tab:jaccard_distmult}
\end{table}

\end{sloppypar}
\end{document}